%% file: bias-mitigation.tex
\documentclass[sigconf]{acmart}
\usepackage{booktabs}
\usepackage{pgfplots}
\usepackage{subcaption}

\usetikzlibrary{arrows, automata, positioning, shapes.geometric, arrows.meta,}
\pgfplotsset{
        compat=1.17,
    }
\AtBeginDocument{%
  }

\setcopyright{acmlicensed}
\copyrightyear{2025}
\acmConference[KDD'25]{The 31st ACM SIGKDD Conference on Knowledge Discovery and Data Mining}{August 03-07, 2025}{Toronto, ON, Canada}
\acmYear{2025}
\acmISBN{XXX-X-XXXX-XXXX-X/25/08}
\acmDOI{10.1145/XXXXXXX.XXXXXXX}
\acmISBN{978-1-4503-XXXX-X/2018/06}

\begin{document}

\title{Bias Mitigation Agent: Optimizing Source Selection for Fair and Balanced Knowledge Retrieval}

\author{Karanbir Singh}
\orcid{0009-0005-2655-6335}
\affiliation{%
  \institution{Salesforce}
  \city{San Francisco}
  \state{California}
  \country{USA}
}
\email{karanbirsingh@salesforce.com}
\authornote{Karanbir Singh is the corresponding author}

\author{Deepak Muppiri}
\orcid{0009-0009-0326-0952}
\affiliation{%
  \institution{Salesforce}
  \city{San Francisco}
  \state{CA}
  \country{USA}
}
\email{dmuppiri@salesforce.com}

\author{William Ngu}
\orcid{0009-0009-0326-0952}
\affiliation{%
  \institution{Salesforce}
  \city{San Francisco}
  \state{CA}
  \country{USA}
}
\email{wngu@salesforce.com}


\begin{abstract}
  Large Language Models (LLMs) have transformed the field of artificial intelligence by unlocking the era of generative applications. Built on top of generative AI capabilities, Agentic AI represents a major shift toward autonomous, goal-driven systems that can reason, retrieve, and act. However, they also inherit the bias present in both internal and external information sources. This significantly affects the fairness and balance of retrieved information, and hence reduces user trust. To address this critical challenge, we introduce a novel Bias Mitigation Agent, a multi-agent system designed to orchestrate the workflow of bias mitigation through specialized agents that optimize the selection of sources to ensure that the retrieved content is both highly relevant and minimally biased to promote fair and balanced knowledge dissemination. The experimental results demonstrate an 81.82\% reduction in bias compared to a baseline naive retrieval strategy.
\end{abstract}


\received{May 26, 2025}

\maketitle

\section{Introduction}
The advent of Large Language Models (LLMs) has undeniably marked a pivotal moment in artificial intelligence, ushering in an era defined by powerful generative capabilities and sophisticated natural language understanding. To take advantage of these capabilities, a variety of prompting techniques have emerged, shaping the way users interact with LLMs. The zero-shot prompting allows LLMs to generate responses without explicit examples provided within the prompt. \citet{kojima2022large} demonstrated that LLMs are not only capable of generating effective responses, but also show intrinsic reasoning abilities when prompted using the zero shot technique. Similarly, Few-Shot prompting involves offering the model a small, explicitly defined set of examples within the prompt, guiding the model's understanding, and improving output accuracy and reasoning capabilities \cite{Brown2020}. To further enhance the capabilities of LLMs, Retrieval-Augmented Generation (RAG) has emerged as an effective extension to traditional prompting techniques. RAG combines retrieval systems with generative language models, enabling the models to access external knowledge sources dynamically during response generation \cite{Lewis2021}.

This technological leap serves as the bedrock for rapidly emerging field of agentic AI, representing a significant paradigm shift towards autonomous systems. Unlike earlier AI applications, these agents are designed not just to respond or generate, but to reason, plan, retrieve information, utilize tools, and execute complex, multi-step tasks to achieve specific goals autonomously. However, as agentic AI systems increasingly rely on LLMs and external information sources to inform their reasoning and actions, they inherit and often amplify their critical vulnerability: bias \cite{Bolukbasi2016}. Bias refers to the consistent imbalance and unjust representation that arises from responses derived from sources that disproportionately privilege or disadvantage specific groups, often mirroring historical or societal inequalities. LLMs are known to capture and reflect the societal biases present in their vast training dataset leading to output that can perpetuate stereotypes related to gender, race, ethnicity, political leaning, and other characteristics \cite{guo2024}. Furthermore, the external knowledge that the agents utilize is itself not neutral. News articles, and other online documents frequently contain skewed perspectives, misinformation, or systemic biases \cite{Pitoura2017}. This propagation of bias directly undermines user trust, compromises system reliability, and poses significant risks of generating harmful or inequitable outcomes \cite{Hu2024}.

While various techniques exist for mitigating bias within the LLMs themselves such as \citet{Zhang2025} presents a model-level debiasing technique using preference optimization algorithm and a debiased preference dataset to address modality bias, where the model over-relies on one modality. These methods often fall short in the dynamic context of Agentic AI workflows. Now a days, the challenge is to actively managing the bias ingested from constantly changing external sources during task execution. Existing agent frameworks often prioritize task completion and information relevance, latency over robust and integrated mechanisms to evaluate and mitigate the bias. This gap highlights a critical need for novel approaches that address bias directly at the point of information retrieval within agentic architectures.

To address this gap, we introduce the Bias Mitigation Agent, a novel multi-agent framework specifically designed to operate within Agentic AI workflows. Our approach automates the bias mitigation process by optimizing the selection of potential information sources prior to ingestion. We propose two techniques for source selection: a zero-shot approach and a few-shot approach, both designed to dynamically assess and mitigate bias during agent operation, thus enhancing fairness, reliability, and overall system trustworthiness. Our results show that this bias reduction is achieved without a corresponding loss in information relevance, showcasing the potential of our approach to promote the development of more responsible, trustworthy, and equitable Agentic AI systems.

The rest of the paper details the related work and architecture of the Bias Mitigation Agent. Also, we present results of comprehensive experiments designed to validate its effectiveness, demonstrating significant quantitative improvements in retrieving unbiased documents across various scenarios compared to baseline agentic retrieval.

\section{Related Work}
In this section, we discuss the existing work that was done to identify and mitigate bias from AI driven systems. These techniques can be categorized based on the stage of the AI lifecycle at which they are implemented: pre-processing, in-processing, and post-processing. In addition to these traditional categories, we also include prompt and agentic bias mitigation approaches to capture emerging work that leverages prompt engineering and autonomous agents to address bias.

\subsection{Pre-processing techniques}
Pre-processing techniques aim to mitigate biases within datasets before they are used for training models, thereby reducing the risk of perpetuating systemic unfairness and thus inherently producing fair models. \citet{Kamiran2011} proposed three data preprocessing techniques: Massaging, Reweighting, and Sampling to address discrimination and mitigate bias in classification tasks. \citet{De_Arteaga_2019}removed gender-related words from a set of biographies which resulted in significant improvement in the fairness of a classifier used to predict corresponding occupations. \citet{Raza2022} introduced Dbias, an open-source Python package designed to detect and mitigate biases in news articles. Dbias pipeline is made up of three core modules: bias detection, bias recognition, and de-biasing. The pipeline ensures that pre-processed data is free of bias, resulting in fairer models during training.

\subsection{In-processing techniques}
While pre-processing techniques focus on data preparation, in-processing approaches tackle bias directly during model training or inference. The idea is to penalize the model if it favors bias and hence it controls the loss function to minimize bias. For example, \citet{Rekabsaz2021} develop AdvBert, a BERT based ranking model that uses adversarial training to simultaneously predict relevance and suppress protected attributes in content retrieved by IR systems. \citet{Jaenich2024} modify the ranking process using policies to ensure that different document categories are ranked fairly and hence improving fairness metrics by 13\% in IR systems. \citet{Singh2019} propose a generic fairness-aware learning-to-rank (LTR) framework using a policy-gradient method to enforce fairness constraints within a listwise LTR setting. Building on this, \citet{Zehlike2020} integrate fairness into listwise LTR by incorporating a regularization term into the model’s utility objective.

\subsection{Post processing techniques}
Post processing introduces fairness after the model or ranking output is generated. \citet{Yang2017} proposed fairness measures for ranked outputs and incorporated these measures into an optimization framework to improve fairness while maintaining accuracy. \citet{Zehlike2017} introduced FA*IR, a post-processing algorithm to ensure group fairness in the retrieved documents by guaranteeing a minimum proportion of protected candidates while maximizing utility in IR systems.

\subsection{Prompting techniques}
The way users interact with LLMs, and the instructions given to the model, can significantly influence bias expression. Prompt engineering techniques aim to guide the model towards fairer outputs by providing specific instructions (e.g., "avoid stereotypes," "provide balanced views"), setting context, or using role-playing prompts. \citet{Huan2023} proposed a novel search strategy based on greedy search to identify the near-optimal prompt to improve the performance of LLM's. \citet{Kamruzzaman2024} explores prompting techniques inspired by dual process theory to reduce social biases in LLM's, techniques such as human and machine like personals, explicit debiasing instructions and chain of though prompting are used to influence the model output to reduce stereotypical responses.

\subsection{Agentic mitigation techniques}
The autonomy and interactive nature of Agentic AI require specific mitigation approaches. Given agents' reliance on external information, strategies focusing on bias-aware information retrieval and source selection are crucial. This can involve integrating bias detectors as tools within the agent's framework to evaluate potential sources before using the information \cite{Singh2025}. \citet{Borah2024} proposes a multi-agent approach, where a metric is developed to assess the presence of bias and then self-reflection and supervised fine-tuning strategies are employed to mitigate bias. \citet{Xu2025} tackles bias mitigation using a multi-objective approach within a multi-agent framework(MOMA) to mitigate bias. Multiple agents are deployed and perform interventions on the bias-related contents of the query.

The approach we present uses agentic mitigation techniques, where multiple agents, such as knowledge, bias detector, source selector, and writer, interact and perform specific tasks to mitigate bias in the output provided to the user without degrading performance. 

\begin{figure*}[!t]
    \centering
    \includegraphics[width=7in]{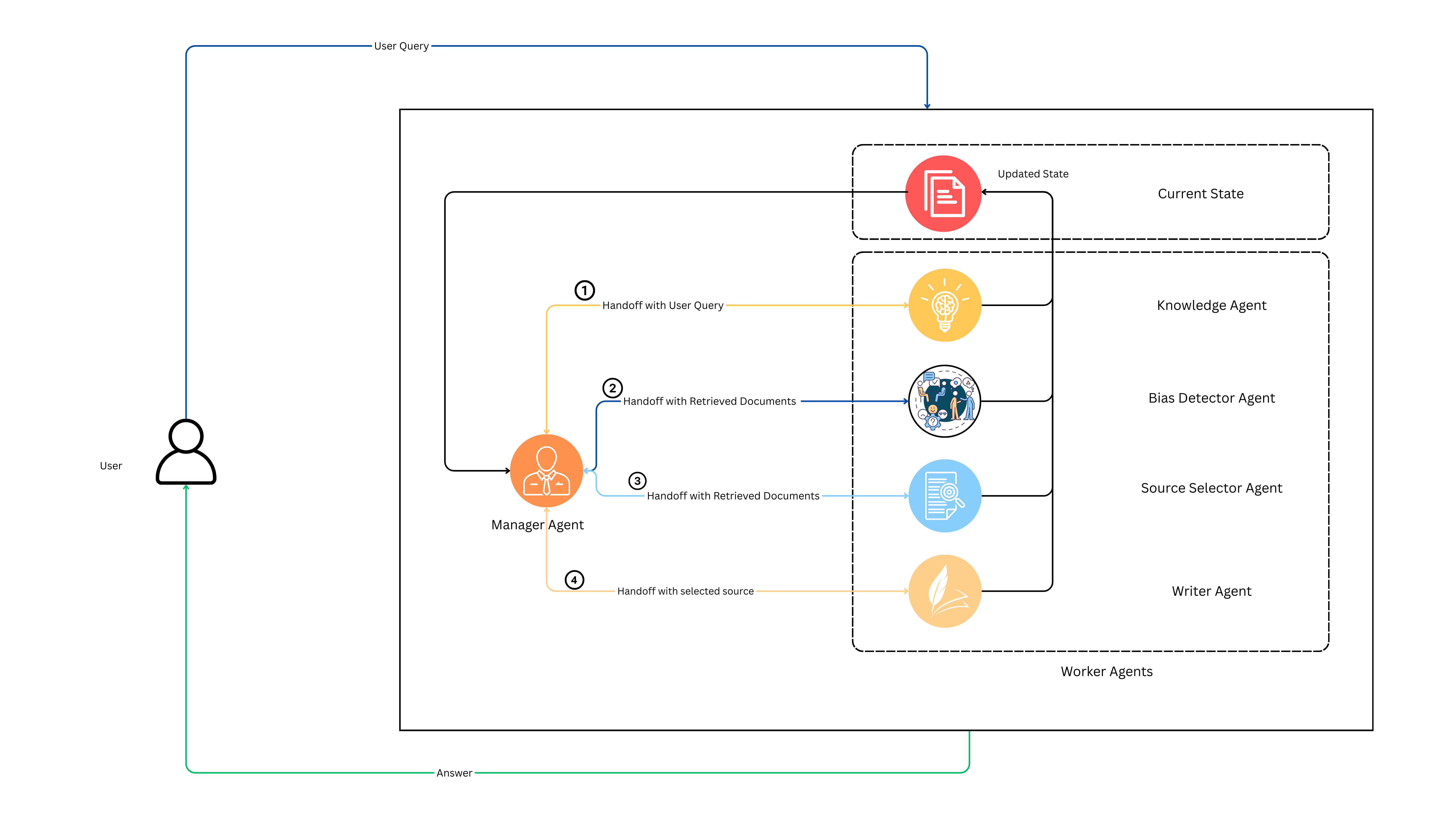}
    \caption{Architecture of a Supervisor-Based Multi-Agent System for Bias Mitigation. The diagram depicts the life cycle of the agent from user's query to final answer coordinated by the manager agent. The Knowledge Agent retrieves documents, the Bias Detection Agent evaluates retrieved documents for bias, and the Source Selector Agent chooses the optimal unbiased sources. Finally, the Writer Agent synthesizes a coherent, unbiased answer, which is presented back to the user.}
    
  \Description{Architecture of a Supervisor-Based Multi-Agent System for Bias Mitigation. The diagram depicts the life cycle of the agent from user's query to final answer coordinated by the manager agent. The Knowledge Agent retrieves documents, the Bias Detection Agent evaluates retrieved documents for bias, and the Source Selector Agent chooses the optimal unbiased sources. Finally, the Writer Agent synthesizes a coherent, unbiased answer, which is presented back to the user.}
  
  \label{fig:architecture}
\end{figure*}

\section{Approach}
This section outlines the architecture and operational design of the Bias Mitigation Framework, a multi-agent system constructed using LangGraph \cite{langgraph} which supports stateful workflows to manage inter-agent communication and control flow. The framework is designed to enhance fairness and transparency in knowledge-retrieval tasks by orchestrating a set of specialized agents through a centralized control mechanism.

At its core, the system consists of a Manager Agent ($M$), a set of Worker Agents ($W$), and a shared state of the system ($\mathcal{S}$). Each worker agent $w_i \in W$ is responsible for a specific task such as document retrieval, bias detection, source selection, etc. The Manager Agent supervises the execution flow, maintains system state, and coordinates decisions based on intermediate outcomes.

The framework supports three operational modes, enabling different strategies for source selection:

\begin{itemize}
    \item \textbf{No Source Selection}: In this baseline mode, the system retrieves the most relevant document on the basis of vector similarity to the user query. The document is then passed to the writer agent without performing any source selection.

    \item \textbf{Zero-Shot}: In this mode, the system retrieves multiple candidate documents and evaluates them based on relevance and bias. The source selector agent makes a decision based solely on these metrics using its parametric knowledge and reasoning capability. This mode provides a lightweight fairness mechanism without requiring any kind of in-context learning.

    \item \textbf{Few-Shot}: This advanced mode leverages labeled examples to guide the source selector agent in making informed decisions. It combines bias and relevance scores with prior demonstrations to achieve more consistent and nuanced selections, especially in domains with subjective or ambiguous content.
\end{itemize}

By supporting these three modes, the framework enables flexible trade-offs between computational efficiency, fairness enforcement, and generalization capability. In the next subsections, the internal state, behavior of the Manager Agent, and the specialized functions of the Worker Agents are defined. 

\subsection{State}
The framework maintains an internal state to facilitate structured decision-making between agents. This state captures the evolving system context, supports transitions, enforces retry logic, and implements guardrails to ensure that the final user response is both relevant and fair. We define the state $\mathcal{S}$ as a tuple, as presented in Equation~\ref{eq:state}:

\begin{equation}
    \mathcal{S} = (\mathcal{C}, \alpha, \kappa, \mu, \varrho)
    \label{eq:state}
\end{equation}

where:
\begin{itemize}
  \item $\mathcal{C} = \{c_1, c_2, \dots, c_n\} \subseteq \mathcal{D}$ is the set of candidate documents retrieved from the vector store, where $\mathcal{D}$ represents the complete corpus of documents.
  \item $\alpha \in \mathcal{C}$ denotes the document selected for final answer generation.
  \item $\kappa \in \mathbb{N}$ is the current retry attempt.
  \item $\mu \in \mathbb{N}$ specifies the maximum number of retries allowed.
  \item $\varrho \in \mathbb{S}$ captures the most recent rejection reason (e.g., all candidates rejected due to bias or low relevance).
\end{itemize}

Each document $d_i \in \mathcal{D}$ is further modeled as a tuple, as defined in Equation~\ref{eq:document}:

\begin{equation}
    d_i = (\chi, \rho, \beta, \gamma)
    \label{eq:document}
\end{equation}

where:
\begin{itemize}
  \item $\chi$ is the textual content of the document.
  \item $\rho \in [0, 1]$ is the relevance score with respect to the user query.
  \item $\beta \in [0, 1]$ is the bias confidence score, which represents the system’s confidence in its bias assessment.
  \item $\gamma \in \{0, 1\}$ is a binary label that indicates whether the document is biased ($\gamma = 1$) or unbiased ($\gamma = 0$).
\end{itemize}

This formalism enables the framework to track a comprehensive and interpretable state across multiple agent interactions and adapt the retrieval process in response to fairness constraints.

\subsection{Manager Agent}
\begin{figure}[!h]
    \centering
    \begin{tikzpicture}
    \node[draw=blue, thick, rounded corners=10pt, inner sep=10pt, text width=8cm, align=left] (box) 
    {
        \textbf{\textless System Prompt with source selection\textgreater} \\
            You are a Supervisor Agent responsible for coordinating multiple specialized agents in a multi-agent system. Your primary goal is to answer user queries using the knowledge provided only and try to minimize bias as much as possible. \\[0.3cm]
            Hand off to the \texttt{knowledge\_agent} to gather information. \\
            Hand off to the \texttt{bias\_detector\_agent} to measure bias inside the retrieved documents. \\
            Hand off to the \texttt{selector} to select a source using relevance and bias scores. \\
            Hand off to the \texttt{writer} to answer the query based on the selected source. \\
            If this is the final answer, return \texttt{\_\_end\_\_} to finish execution.
    };
    
\end{tikzpicture}
    \caption{System prompt used for manager when source selection is enabled.}
    \label{fig:prompt-with-selection}
\end{figure}

\begin{figure}[!h]
    \centering
    \begin{tikzpicture}
    \node[draw=blue, thick, rounded corners=10pt, inner sep=10pt, text width=8cm, align=left] (box) 
    {
        \textbf{\textless System Prompt without source selection\textgreater} \\
        You are a Supervisor Agent responsible for coordinating multiple specialized agents in a multi-agent system. Your primary goal is to answer user queries using the knowledge provided only and try to minimize bias as much as possible. \\[0.3cm]
        Hand off to the \texttt{knowledge\_agent} to gather source candidates. \\
        Hand off to the \texttt{bias\_detector\_agent} to measure bias of the retrieved document. \\
        Hand off to the \texttt{writer} to answer the query based on the selected source. \\
        If this is the final answer, return \texttt{\_\_end\_\_} to finish execution. \\
    };
\end{tikzpicture}
    \caption{System prompt used for manager when source selection is disabled.}
    \label{fig:prompt-without-selection}
\end{figure}
The manager agent is the coordinator within the Bias Mitigation Framework. It is responsible for maintaining the current state of the system, directing the sequence of agent invocations, and enforcing retry policies in the event of retrieval/selection failures. The agent is guided by system-level prompts, which vary depending on the execution mode. These prompts are illustrated in Figures ~\ref{fig:prompt-with-selection} and~\ref{fig:prompt-without-selection}.

\subsection{Worker Agents}
The Worker Agents are specialized components of the Bias Mitigation Framework.  Unlike the Manager Agent, which controls the orchestration and high-level control flow, these agents focus on specific tasks. The framework consists of the following worker agents: 

\subsubsection{Knowledge Agent}
This agent is implemented as a tool-calling agent that interfaces with ChromaDB \cite{chromadb}, which acts as a retriever. Its primary responsibility is to fetch the top-$k$ documents from the corpus based on vector similarity to the user query $q$. The agent operates differently depending on whether source selection is enabled.

In case of no source selection mode, the agent retrieves a single document \( d \in \mathcal{D} \) that maximizes relevance and is automatically chosen as the selected source \(\alpha\) without using any sophisticated process to mitigate bias, reflecting the typical behavior of the current LLM-based information retrieval systems in production as presented in the following equation~\ref{eq:knowlegde_agent}:

\begin{equation}
\alpha = d = \arg\max_{d_i \in \mathcal{D}} \rho_i
\label{eq:knowlegde_agent}
\end{equation}

Here, \( \rho_i \) denotes the relevance score of the document \( d_i \).

When source selection is enabled, the agent retrieves a set of candidate documents $\mathcal{C} \subseteq \mathcal{D}$ to allow downstream agents to assess both relevance and bias. If all candidate documents \( c_i \in \mathcal{C} \) are rejected due to high bias or low relevance, then the system retries to retrieve ideal candidates \(\mathcal{C} \). In the retry phase, the agent performs query expansion, transforming the original query $q$ into an improved query $q'$ based on the rejection reason $\varrho$. The new query $q'$ is embedded as \( v_{q'} \), and a new set of candidate documents is selected. The resulting set of candidates is passed on to downstream agents for further evaluation.

\subsubsection{Bias Detection Agent}
This Agent is responsible for evaluating the presence and severity of bias at the source level. When source selection is enabled, the manager forwards the candidate set $\mathcal{C}$ retrieved by the knowledge agent to the agent. Each candidate document \( c_i \in \mathcal{C} \) is then analyzed using a pre-trained text classification model called Dbias \cite{Raza2022}. For each candidate document, the agent assigns the following:
\begin{itemize}
    \item A bias confidence score ($\beta_i \in [0,1]$), which quantifies the system confidence in the detected bias.
    \item A binary label ($\gamma_i \in \{0,1\}$), where $\gamma_i = 1$ indicates that the document is biased, and $\gamma_i = 0$ indicates that it is unbiased.
\end{itemize}

These values are then used to update the current state $\mathcal{S}$. In the case of no source selection mode, the agent operates on the selected document $\alpha$, applying the same analysis pipeline. This ensures that even in the absence of comparative selection, the system retains awareness of potential bias in the final chosen source.

\subsubsection{Source Selection Agent}
The agent is responsible for identifying the most suitable document from the candidate set $\mathcal{C}$ by evaluating both the relevance and the bias metrics. This agent is invoked only when the system operates in modes that enable source selection, specifically zero-shot and few-shot.

\paragraph{Zero Shot} In this mode, the agent applies a rule-based selection strategy and uses the parametric knowledge and reasoning ability of the underlying model to select a suitable source $\alpha$ to answer the user's query $q$. During the \textit{first attempt}, it adheres to strict selection criteria as defined in equation~\ref{eq:first_attempt} where only documents with $\gamma_i = 0$ (unbiased) and $\beta_i \geq 0.7$ (high confidence in the determination) are considered. Among these, the document with the highest relevance score $\rho_i$ is selected.

\begin{equation}
\alpha = \arg\max_{c_i \in \mathcal{C}'} \rho_i, \quad \text{where } \mathcal{C}' = \{c_i \in \mathcal{C} \mid \gamma_i = 0 \land \beta_i \geq 0.7\}
\label{eq:first_attempt}
\end{equation}

If no candidate meets the selection criteria, the agent updates the current state $\mathcal{S}$ with the rejection reason $\varrho$ and the entire retrieval and selection process is executed again by the manager agent. If the system reaches its final attempt, the agent applies relaxed selection rules. This allows the system to still generate an answer even under constrained document conditions.

\paragraph{Few Shot} In this mode, the agent uses in-context examples to guide its decision making. It is provided with a set of labeled instances that illustrate how to select the optimal document based on combinations of $(\beta_i, \gamma_i, \rho_i)$ values. These examples encode decision patterns that help the agent generalize the source selection logic beyond simple thresholding mechanisms.

Given a set of candidates $\mathcal{C}$, the agent evaluates each candidate $c_i$ based on its similarity to previous examples and selects the most appropriate document that meets the dual criteria of high relevance and minimal bias. Formally:

\begin{equation}
\alpha = \arg\max_{c_i \in \mathcal{C}} f_{\text{few-shot}}(\beta_i, \gamma_i, \rho_i)
\end{equation}

where $f_{\text{few-shot}}$ is a learned or example-conditioned scoring function implicitly encoded via prompt demonstrations.

Similarly to the zero-shot mode, if no candidate meets the selection criteria, then the system attempts to retry and pick the suitable candidate as the selected source $\alpha$. 

\subsubsection{Writer Agent}
The agent is responsible for generating the final response to the user's query $q$. It takes the selected document $\alpha$, as determined by the system, and synthesizes a coherent and contextually grounded response. It is provided with the original user query $q$ and the selected source document $\alpha$ as part of a structured system prompt. The agent is explicitly instructed to rely only on the content of the provided source for its answer generation. This helps ensure factual accuracy and reduces bias by limiting the response to the selected source. The writer agent marks the final stage of the pipeline. Upon generating a satisfactory response, the manager agent terminates the execution and returns the output to the user.

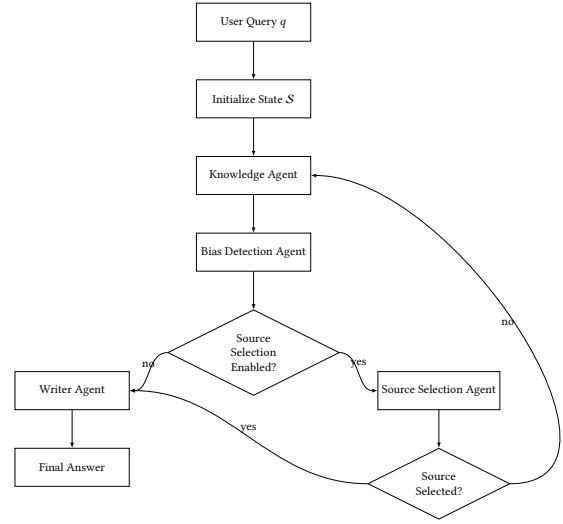
\begin{figure}[!t]
\centering
\resizebox{0.48\textwidth}{!}{%
\begin{tikzpicture}[node distance=1cm and 1cm]
    \tikzstyle{diam} = [diamond, aspect=2, draw=black, text width=6em, text centered ]
    \tikzstyle{rect} = [rectangle, minimum width=3cm, minimum height=1cm, text centered, draw=black]
    \tikzstyle{line} = [draw, -latex]

    \node [rect] (query) {User Query $q$} ;
    \node [rect, below=of query] (init) {Initialize State $\mathcal{S}$};
    \node [rect, below=of init] (knowledge) {Knowledge Agent};
    \node [rect, below=of knowledge] (bias_detect) {Bias Detection Agent};
    \node [diam, below=of bias_detect] (mode) {Source Selection Enabled?};
    \node [rect, right=of mode, yshift=-1cm] (selector) {Source Selection Agent};
    \node [rect, left=of mode, yshift=-1cm] (writer) {Writer Agent};
    \node [rect, below=of writer] (final) {Final Answer};
    \node [diam, below=of selector] (source_selected) {Source Selected?};

    \path [line] (query) -- (init);
    \path [line] (init) -- (knowledge);
    \path [line] (knowledge) -- (bias_detect);
    \path [line] (bias_detect) -- (mode);
    \path [line] (writer) -- (final);
    \path [line] (mode) to[out=0, in=180] node[above] {yes} (selector) ;
    \path [line] (mode) to[out=180, in=0] node[above] {no} (writer) ;
    \path [line] (selector) -- (source_selected);
    \path [line] (source_selected) to[out=180, in=0] node[above] {yes} (writer) ;
    \path [line] (source_selected) to[out=0, in=0] node[above left] {no} (knowledge);
\end{tikzpicture}
}
\caption{Execution flow of the Bias Mitigation Framework across all operational modes}
\label{fig:execution-flow-diagram}
\end{figure}

Together, these Worker Agents form a tightly integrated and modular pipeline within the Bias Mitigation Framework. Each agent is designed to fulfill a distinct and well-scoped responsibility, enabling separation of concerns and ease of extension. By delegating complex tasks such as retrieval, bias evaluation, source selection, and response generation to specialized agents, the system ensures robustness, adaptability, and transparency across diverse user queries and fairness constraints. Figure~\ref{fig:execution-flow-diagram} elaborates on the end-to-end execution flow, highlighting how these agents interact in different operating modes to ensure reliable and bias minimized answer to the user's query $q$.

\section{Experimentation}
\input{plots_tex/table-lat-bi-rel}
This section evaluates the performance of the Bias Mitigation Agent in its ability to reduce bias while maintaining relevance to the original user query. The agent uses annotated news articles sourced from the MBIC \cite{spinde2021mbic} and BABE datasets \cite{Spinde2021f} to evaluate its bias detection and mitigation capabilities. We curated \textit{112 queries} to validate the bias mitigation capabilities of the agent. To conduct these experiments, we utilized OpenAI's GPT series models such as GPT-4o-mini, GPT-4.1, and GPT-4.1-mini as the underlying reasoning engines. The following sections provide a detailed analysis of each approach and present their respective outcomes.

\subsection{Baseline: No Source Selection}
No Source Selection operates by selecting the most relevant source document, without bias filtering. As shown in Table \ref{table:combined_llm_metrics} has the fastest response time among the methods, averaging 12.06 seconds per query ($\pm$ 2.45) among all models and modes. It achieved a relevance score of 0.181 ($\pm$ 0.131) when using GPT-4.1-mini. However, this speed and relevance came at the expense of fairness, with 49.11\% of the outputs labeled biased using GPT-4o-mini, 56.25\% using GPT-4.1, and 52.68\% using GPT-4.1-mini as shown in Figure \ref{fig:bias_reduction}. The bias classifier's average confidence score for bias in this mode is 0.8402 ($\pm$ 0.1464), 0.858 ($\pm$ 0.145), and 0.833 ($\pm$ 0.163) using GPT-4o-mini, GPT-4.1, GPT-4.1-mini, respectively.

\input{plots_tex/bias_reduction_new}
\subsection{Zero-Shot}
In case of GPT-4o-mini, the zero-shot mode achieved the lowest bias rate at 8.929\%, significantly outperforming the baseline by approximately 81.82\% as presented in Figure \ref{fig:bias_reduction}. The beat average relevance score was 0.366 ($\pm$ 0.367), which is even better than the baseline model in terms of utility of the answers to the queries between all models and modes. However, the zero-shot mode incurred the highest latency, averaging 41.38 seconds ($\pm$ 13.89) for GPT-4o-mini, 28.66 seconds ($\pm$ 13.41) for GPT-4.1, and 35.45 seconds ($\pm$ 15.98) for GPT-4.1-mini, as given in Table \ref{table:combined_llm_metrics}. It also has taken multiple hops in 70.54\%, 21.43\%, and 26.79\% of queries using GPT-4o-mini, GPT-4.1, and GPT-4.1-mini as depicted in Figure \ref{fig:retry_rates} respectively. The high retry rate suggests that it adopts a more cautious and iterative selection strategy due to the lack of in-context information. Lastly, its average bias confidence was 0.805 ($\pm$ 0.1431) using GPT-4o-mini, 0.841 ($\pm$ 0.130) using GPT-4.1, and 0.837 ($\pm$ 0.123) using GPT-4.1-mini.

\subsection{Few-Shot}
The few-shot mode utilizes in-context examples to improve selection accuracy. Figure \ref{fig:bias_reduction} shows that in case of GPT-4o-mini, it achieved a bias rate of 14.3\%, demonstrating a substantial improvement over the baseline mode by 69.48\%. Also, while using GPT-4.1 and GPT-4.1-mini, it generated bias results only 17.86\% and 23.21\% of the time which is the lowest when using these models as reasoning engines. As mentioned in Table \ref{table:combined_llm_metrics}, the average relevance score of 0.236 ($\pm$ 0.251) was achieved, which is higher than the baseline mode for all models. The major improvement came with the average latency which is 37.79 seconds ($\pm$ 11.40), 26.93 seconds ($\pm$ 11.99), and 30.58 seconds ($\pm$ 13.63) using GPT-4o-mini, GPT-4.1, GPT-4.1-mini; therefore, it is slightly faster than the zero-shot mode while all the time, as shown in Table \ref{table:combined_llm_metrics}. Also, Figure \ref{fig:retry_rates} shows that it is more decisive as it only retried 29.46\%, 18.75\%, and 16.07\% of the time using GPT-4o-mini, GPT-4.1, and GPT-4.1-mini.

\input{plots_tex/retry_rates}

In summary, when comparing both models, GPT-4.1-mini consistently outperformed GPT-4o-mini, GPT-4.1 in relevance, achieving the highest average relevance score of 0.366 ($\pm$ 0.367) in zero-shot mode which supports its superior ability to generate contextually aligned responses. On the other hand, in terms of bias reduction, GPT-4o-mini with zero-shot mode achieved the lowest bias rate overall at 8.93\%, although with increased latency and variability.

\section{Conclusion}

In this paper, we introduce the Bias Mitigation Agent, a novel multi-agent framework designed to enhance fairness and trust in agentic information retrieval systems by optimizing the source selection process. By using specialized agents for knowledge retrieval, bias detection, and source selection in a supervisor-based architecture, our system enables dynamic and transparent decision-making for mitigating bias in real time.

We evaluated three operational modes, No Source Selection, zero-shot and few-shot across \textit{112 queries} using annotated datasets using GPT-4o-mini, GPT-4.1, GPT-4.1-mini. The results showed that GPT-4o-mini achieved the lowest overall bias rate (8.93\%) in the zero-shot mode, while GPT-4.1-mini consistently outperformed in relevance, with a maximum average relevance score (0.366 $\pm$ 0.367) in the zero-shot mode.

The modular design of the framework allows for extensibility, making it adaptable for future integrations with more advanced bias detectors, domain-specific retrievers, and additional modalities. As the field of Agentic AI continues to evolve, our work highlights the critical role of optimizing workflows in responsible knowledge retrieval and paves the way for more equitable and trustworthy AI systems.

Future directions include fine-tuning bias scoring functions with human feedback, exploring reinforcement learning for adaptive source selection, and extending the framework to handle multimodal inputs such as images and audio.

\bibliographystyle{ACM-Reference-Format}
\bibliography{bias-mitigation}
\end{document}

%% file: plots_tex/table-lat-bi-rel.tex
\begin{table*}[t]
\centering
\resizebox{\textwidth}{!}{%
\begin{tabular}{llcccccccccc}
\toprule
\textbf{Reasoner Model} & \textbf{Mode} & \textbf{Rel Min} & \textbf{Rel Max} & \textbf{Rel Avg ± Std} & \textbf{Bias Min} & \textbf{Bias Max} & \textbf{Bias Avg ± Std} & \textbf{Lat Min} & \textbf{Lat Max} & \textbf{Lat Avg ± Std} \\
\midrule
4o-mini & No Source Selection & -0.058 & 0.426 & 0.169 ± 0.092 & 0.531 & 0.995 & 0.840 ± 0.146 & 8.77 & 39.00 & 17.88 ± 4.59 \\
4o-mini & Zero-Shot           &  0.006 & 0.402 & 0.157 ± 0.078 & 0.515 & 0.995 & 0.806 ± 0.143 & 16.68 & 69.51 & 41.38 ± 13.89 \\
4o-mini & Few-Shot            & -0.027 & 0.395 & 0.150 ± 0.084 & 0.521 & 0.995 & 0.813 ± 0.140 & 23.92 & 64.20 & 37.78 ± 11.40 \\
\midrule
4.1 & No Source Selection     & -0.058 & 0.464 & 0.172 ± 0.096 & 0.531 & 0.995 & 0.858 ± 0.145 & 8.97 & 20.47 & 12.06 ± 2.45 \\
4.1 & Zero-Shot               & -0.005 & 0.971 & 0.197 ± 0.192 & 0.524 & 0.995 & 0.841 ± 0.130 & 11.25 & 81.64 & 28.66 ± 13.41 \\
4.1 & Few-Shot                & -0.058 & 0.978 & 0.171 ± 0.140 & 0.515 & 0.995 & 0.801 ± 0.146 & 11.47 & 65.37 & 26.93 ± 11.99 \\
\midrule
4.1-mini & No Source Selection & -0.058 & 0.907 & 0.181 ± 0.131 & 0.099 & 0.995 & 0.833 ± 0.163 & 11.02 & 43.77 & 17.12 ± 6.74 \\
4.1-mini & Zero-Shot           & -0.058 & 0.950 & 0.366 ± 0.367 & 0.531 & 0.995 & 0.837 ± 0.123 & 13.54 & 76.09 & 35.45 ± 15.98 \\
4.1-mini & Few-Shot            & -0.058 & 0.950 & 0.236 ± 0.251 & 0.501 & 0.995 & 0.829 ± 0.143 & 13.15 & 78.67 & 30.58 ± 13.63 \\
\bottomrule
\end{tabular}%
}
\caption{Summary of relevance, bias confidence, and latency for each source selection method across GPT-4o-mini, GPT-4.1, and GPT-4.1-mini. A divider separates rows from different models.}
\label{table:combined_llm_metrics}
\end{table*}

%% file: plots_tex/bias_reduction_new.tex
\begin{figure}[h]
\centering
\begin{tikzpicture}
\begin{axis}[
    ybar=0pt,
    bar width=14pt,
    width=\linewidth,
    height=6cm,
    ymin=0, ymax=70,
    ylabel={Bias Rate (\%)},
    symbolic x coords={No Source Selection, Zero-Shot, Few-Shot},
    xtick=data,
    xticklabel style={font=\small},
    legend style={
        at={(0.5,-0.25)},
        anchor=north,
        legend columns=3,
        font=\footnotesize
    },
    nodes near coords,
    every node near coord/.append style={
        font=\scriptsize, yshift=5pt, color=black
    },
    enlarge x limits=0.15,
    grid=major,
    axis lines*=box,
    label style={font=\small}
]

\addplot+[fill=gray!40] coordinates {
    (No Source Selection,49.11)
    (Zero-Shot,8.93)
    (Few-Shot,14.29)
};
\addlegendentry{GPT-4o-mini}

\addplot+[fill=blue!50] coordinates {
    (No Source Selection,56.25)
    (Zero-Shot,19.64)
    (Few-Shot,17.86)
};
\addlegendentry{GPT-4.1}

\addplot+[fill=orange!60] coordinates {
    (No Source Selection,52.68)
    (Zero-Shot,27.68)
    (Few-Shot,23.21)
};
\addlegendentry{GPT-4.1-mini}

\end{axis}
\end{tikzpicture}
\vspace{-1em}
\caption{Bias rate comparison for each source selection method across GPT-4o-mini, GPT-4.1, and GPT-4.1-mini. GPT-4o-mini performs strongest on bias mitigation overall, especially under Zero-Shot prompting.}
\label{fig:bias_reduction}
\end{figure}
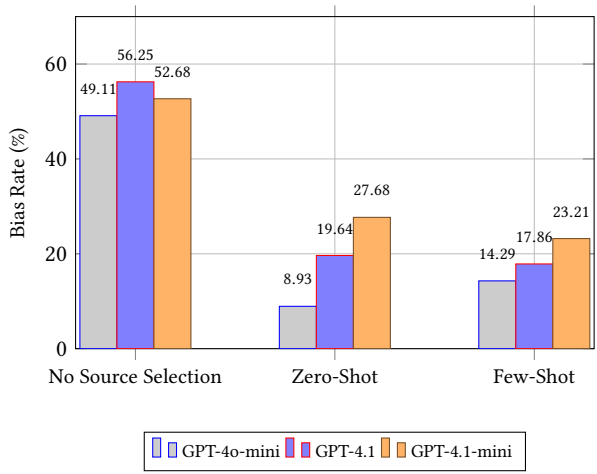

%% file: plots_tex/retry_rates.tex
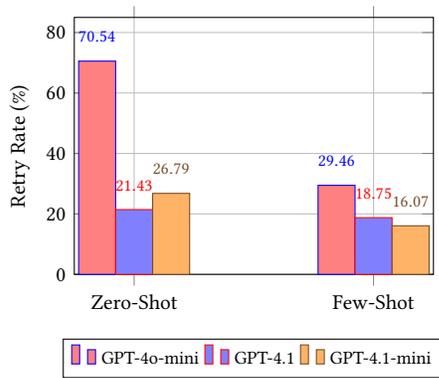
\begin{figure}[h]
    \centering
    \begin{tikzpicture}
    \begin{axis}[
        width=0.75\linewidth,
        height=5cm,
        ybar=0pt,
        bar width=14pt,
        ymin=0,
        ymax=85,
        ylabel={Retry Rate (\%)},
        symbolic x coords={Zero-Shot, Few-Shot},
        xtick=data,
        enlarge x limits=0.25,
        nodes near coords,
        every node near coord/.append style={font=\scriptsize, yshift=4pt},
        grid=major,
        legend style={
            at={(0.5,-0.25)},
            anchor=north,
            legend columns=3,
            font=\footnotesize
        },
        tick label style={font=\small},
        label style={font=\small}
    ]

    \addplot+[fill=red!50] coordinates {
        (Zero-Shot,70.54)
        (Few-Shot,29.46)
    };
    \addlegendentry{GPT-4o-mini}

    \addplot+[fill=blue!50] coordinates {
        (Zero-Shot,21.43)
        (Few-Shot,18.75)
    };
    \addlegendentry{GPT-4.1}

    \addplot+[fill=orange!60] coordinates {
        (Zero-Shot,26.79)
        (Few-Shot,16.07)
    };
    \addlegendentry{GPT-4.1-mini}

    \end{axis}
    \end{tikzpicture}
    \caption{Retry rates for Zero-Shot and Few-Shot selectors across GPT-4o-mini, GPT-4.1, and GPT-4.1-mini. GPT-4o-mini shows the highest retry activity, especially under Zero-Shot, while GPT-4.1-mini exhibits a more balanced retry profile.}
    \label{fig:retry_rates}
\end{figure}

%% file: bias-mitigation.bbl

\begin{thebibliography}{26}


\ifx \showCODEN    \undefined \def \showCODEN     #1{\unskip}     \fi
\ifx \showISBNx    \undefined \def \showISBNx     #1{\unskip}     \fi
\ifx \showISBNxiii \undefined \def \showISBNxiii  #1{\unskip}     \fi
\ifx \showISSN     \undefined \def \showISSN      #1{\unskip}     \fi
\ifx \showLCCN     \undefined \def \showLCCN      #1{\unskip}     \fi
\ifx \shownote     \undefined \def \shownote      #1{#1}          \fi
\ifx \showarticletitle \undefined \def \showarticletitle #1{#1}   \fi
\ifx \showURL      \undefined \def \showURL       {\relax}        \fi
\providecommand\bibfield[2]{#2}
\providecommand\bibinfo[2]{#2}
\providecommand\natexlab[1]{#1}
\providecommand\showeprint[2][]{arXiv:#2}

\bibitem[Bolukbasi et~al\mbox{.}(2016)]%
        {Bolukbasi2016}
\bibfield{author}{\bibinfo{person}{Tolga Bolukbasi}, \bibinfo{person}{Kai-Wei Chang}, \bibinfo{person}{James Zou}, \bibinfo{person}{Venkatesh Saligrama}, {and} \bibinfo{person}{Adam Kalai}.} \bibinfo{year}{2016}\natexlab{}.
\newblock \showarticletitle{Man is to Computer Programmer as Woman is to Homemaker? Debiasing Word Embeddings}.
\newblock \bibinfo{journal}{\emph{arXiv}} (\bibinfo{year}{2016}).
\newblock
\showeprint[arxiv]{1607.06520}~[cs.CL]
\urldef\tempurl%
\url{https://arxiv.org/abs/1607.06520}
\showURL{%
\tempurl}


\bibitem[Borah and Mihalcea(2024)]%
        {Borah2024}
\bibfield{author}{\bibinfo{person}{Angana Borah} {and} \bibinfo{person}{Rada Mihalcea}.} \bibinfo{year}{2024}\natexlab{}.
\newblock \bibinfo{title}{Towards Implicit Bias Detection and Mitigation in Multi-Agent LLM Interactions}.
\newblock
\showeprint[arxiv]{2410.02584}~[cs.CL]
\urldef\tempurl%
\url{https://arxiv.org/abs/2410.02584}
\showURL{%
\tempurl}


\bibitem[Brown et~al\mbox{.}(2020)]%
        {Brown2020}
\bibfield{author}{\bibinfo{person}{Tom~B. Brown}, \bibinfo{person}{Benjamin Mann}, \bibinfo{person}{Nick Ryder}, \bibinfo{person}{Melanie Subbiah}, \bibinfo{person}{Jared Kaplan}, \bibinfo{person}{Prafulla Dhariwal}, \bibinfo{person}{Arvind Neelakantan}, \bibinfo{person}{Pranav Shyam}, \bibinfo{person}{Girish Sastry}, \bibinfo{person}{Amanda Askell}, \bibinfo{person}{Sandhini Agarwal}, \bibinfo{person}{Ariel Herbert-Voss}, \bibinfo{person}{Gretchen Krueger}, \bibinfo{person}{Tom Henighan}, \bibinfo{person}{Rewon Child}, \bibinfo{person}{Aditya Ramesh}, \bibinfo{person}{Daniel~M. Ziegler}, \bibinfo{person}{Jeffrey Wu}, \bibinfo{person}{Clemens Winter}, \bibinfo{person}{Christopher Hesse}, \bibinfo{person}{Mark Chen}, \bibinfo{person}{Eric Sigler}, \bibinfo{person}{Mateusz Litwin}, \bibinfo{person}{Scott Gray}, \bibinfo{person}{Benjamin Chess}, \bibinfo{person}{Jack Clark}, \bibinfo{person}{Christopher Berner}, \bibinfo{person}{Sam McCandlish}, \bibinfo{person}{Alec Radford}, \bibinfo{person}{Ilya Sutskever},
  {and} \bibinfo{person}{Dario Amodei}.} \bibinfo{year}{2020}\natexlab{}.
\newblock \bibinfo{title}{Language Models are Few-Shot Learners}.
\newblock
\showeprint[arxiv]{2005.14165}~[cs.CL]
\urldef\tempurl%
\url{https://arxiv.org/abs/2005.14165}
\showURL{%
\tempurl}


\bibitem[Chroma(2022)]%
        {chromadb}
\bibfield{author}{\bibinfo{person}{Chroma}.} \bibinfo{year}{2022}\natexlab{}.
\newblock \bibinfo{title}{Chroma: The open-source AI application database}.
\newblock
\urldef\tempurl%
\url{https://www.trychroma.com/}
\showURL{%
\tempurl}
\newblock
\shownote{Accessed: May 2025}.


\bibitem[De-Arteaga et~al\mbox{.}(2019)]%
        {De_Arteaga_2019}
\bibfield{author}{\bibinfo{person}{Maria De-Arteaga}, \bibinfo{person}{Alexey Romanov}, \bibinfo{person}{Hanna Wallach}, \bibinfo{person}{Jennifer Chayes}, \bibinfo{person}{Christian Borgs}, \bibinfo{person}{Alexandra Chouldechova}, \bibinfo{person}{Sahin Geyik}, \bibinfo{person}{Krishnaram Kenthapadi}, {and} \bibinfo{person}{Adam~Tauman Kalai}.} \bibinfo{year}{2019}\natexlab{}.
\newblock \showarticletitle{Bias in Bios: A Case Study of Semantic Representation Bias in a High-Stakes Setting}. In \bibinfo{booktitle}{\emph{Proceedings of the Conference on Fairness, Accountability, and Transparency}} \emph{(\bibinfo{series}{FAT* ’19})}. \bibinfo{publisher}{ACM}, \bibinfo{pages}{120–128}.
\newblock
\href{https://doi.org/10.1145/3287560.3287572}{doi:\nolinkurl{10.1145/3287560.3287572}}


\bibitem[Guo et~al\mbox{.}(2024)]%
        {guo2024}
\bibfield{author}{\bibinfo{person}{Yufei Guo}, \bibinfo{person}{Muzhe Guo}, \bibinfo{person}{Juntao Su}, \bibinfo{person}{Zhou Yang}, \bibinfo{person}{Mengqiu Zhu}, \bibinfo{person}{Hongfei Li}, \bibinfo{person}{Mengyang Qiu}, {and} \bibinfo{person}{Shuo~Shuo Liu}.} \bibinfo{year}{2024}\natexlab{}.
\newblock \bibinfo{title}{Bias in Large Language Models: Origin, Evaluation, and Mitigation}.
\newblock
\showeprint[arxiv]{2411.10915}~[cs.CL]
\urldef\tempurl%
\url{https://arxiv.org/abs/2411.10915}
\showURL{%
\tempurl}


\bibitem[Hu et~al\mbox{.}(2024)]%
        {Hu2024}
\bibfield{author}{\bibinfo{person}{Mengxuan Hu}, \bibinfo{person}{Hongyi Wu}, \bibinfo{person}{Zihan Guan}, \bibinfo{person}{Ronghang Zhu}, \bibinfo{person}{Dongliang Guo}, \bibinfo{person}{Daiqing Qi}, {and} \bibinfo{person}{Sheng Li}.} \bibinfo{year}{2024}\natexlab{}.
\newblock \bibinfo{title}{No Free Lunch: Retrieval-Augmented Generation Undermines Fairness in LLMs, Even for Vigilant Users}.
\newblock
\showeprint[arxiv]{2410.07589}~[cs.IR]
\urldef\tempurl%
\url{https://arxiv.org/abs/2410.07589}
\showURL{%
\tempurl}


\bibitem[Inc(2023)]%
        {langgraph}
\bibfield{author}{\bibinfo{person}{LangChain Inc}.} \bibinfo{year}{2023}\natexlab{}.
\newblock \bibinfo{title}{LangGraph: A Library for Building Multi-Agent Workflows with LLMs}.
\newblock
\urldef\tempurl%
\url{https://github.com/langchain-ai/langgraph}
\showURL{%
\tempurl}
\newblock
\shownote{Accessed: May 2025}.


\bibitem[Jaenich et~al\mbox{.}(2024)]%
        {Jaenich2024}
\bibfield{author}{\bibinfo{person}{Thomas Jaenich}, \bibinfo{person}{Graham McDonald}, {and} \bibinfo{person}{Iadh Ounis}.} \bibinfo{year}{2024}\natexlab{}.
\newblock \showarticletitle{Fairness-Aware Exposure Allocation via Adaptive Reranking}. In \bibinfo{booktitle}{\emph{Proceedings of the 47th International ACM SIGIR Conference on Research and Development in Information Retrieval (SIGIR '24)}}. \bibinfo{publisher}{Association for Computing Machinery}, \bibinfo{address}{Washington, DC, USA}, \bibinfo{pages}{1504--1513}.
\newblock
\href{https://doi.org/10.1145/3626772.3657794}{doi:\nolinkurl{10.1145/3626772.3657794}}


\bibitem[Kamiran and Calders(2011)]%
        {Kamiran2011}
\bibfield{author}{\bibinfo{person}{Faisal Kamiran} {and} \bibinfo{person}{Toon Calders}.} \bibinfo{year}{2011}\natexlab{}.
\newblock \showarticletitle{Data Pre-Processing Techniques for Classification without Discrimination.}
\newblock \bibinfo{journal}{\emph{Knowledge and Information Systems}} (\bibinfo{year}{2011}).
\newblock


\bibitem[Kamruzzaman and Kim(2024)]%
        {Kamruzzaman2024}
\bibfield{author}{\bibinfo{person}{Mahammed Kamruzzaman} {and} \bibinfo{person}{Gene~Louis Kim}.} \bibinfo{year}{2024}\natexlab{}.
\newblock \bibinfo{title}{Prompting Techniques for Reducing Social Bias in LLMs through System 1 and System 2 Cognitive Processes}.
\newblock
\showeprint[arxiv]{2404.17218}~[cs.CL]
\urldef\tempurl%
\url{https://arxiv.org/abs/2404.17218}
\showURL{%
\tempurl}


\bibitem[Kojima et~al\mbox{.}(2022)]%
        {kojima2022large}
\bibfield{author}{\bibinfo{person}{Takeshi Kojima}, \bibinfo{person}{Shixiang~Shane Gu}, \bibinfo{person}{Machel Reid}, \bibinfo{person}{Yutaka Matsuo}, {and} \bibinfo{person}{Yusuke Iwasawa}.} \bibinfo{year}{2022}\natexlab{}.
\newblock \showarticletitle{Large language models are zero-shot reasoners}.
\newblock \bibinfo{journal}{\emph{Advances in neural information processing systems}}  \bibinfo{volume}{35} (\bibinfo{year}{2022}), \bibinfo{pages}{22199--22213}.
\newblock


\bibitem[Lewis et~al\mbox{.}(2020)]%
        {Lewis2021}
\bibfield{author}{\bibinfo{person}{Patrick Lewis}, \bibinfo{person}{Ethan Perez}, \bibinfo{person}{Aleksandra Piktus}, \bibinfo{person}{Fabio Petroni}, \bibinfo{person}{Vladimir Karpukhin}, \bibinfo{person}{Naman Goyal}, \bibinfo{person}{Heinrich K{\"u}ttler}, \bibinfo{person}{Mike Lewis}, \bibinfo{person}{Wen-tau Yih}, \bibinfo{person}{Tim~Rocktäschel Rockt{\"a}schel}, \bibinfo{person}{Sebastian Riedel}, {and} \bibinfo{person}{Douwe Kiela}.} \bibinfo{year}{2020}\natexlab{}.
\newblock \showarticletitle{Retrieval-augmented generation for knowledge-intensive nlp tasks}.
\newblock \bibinfo{journal}{\emph{Advances in Neural Information Processing Systems}}  \bibinfo{volume}{33} (\bibinfo{year}{2020}), \bibinfo{pages}{9459--9474}.
\newblock


\bibitem[Ma et~al\mbox{.}(2023)]%
        {Huan2023}
\bibfield{author}{\bibinfo{person}{Huan Ma}, \bibinfo{person}{Changqing Zhang}, \bibinfo{person}{Yatao Bian}, \bibinfo{person}{Lemao Liu}, \bibinfo{person}{Zhirui Zhang}, \bibinfo{person}{Peilin Zhao}, \bibinfo{person}{Shu Zhang}, \bibinfo{person}{Huazhu Fu}, \bibinfo{person}{Qinghua Hu}, {and} \bibinfo{person}{Bingzhe Wu}.} \bibinfo{year}{2023}\natexlab{}.
\newblock \bibinfo{title}{Fairness-guided Few-shot Prompting for Large Language Models}.
\newblock
\showeprint[arxiv]{2303.13217}~[cs.CL]
\urldef\tempurl%
\url{https://arxiv.org/abs/2303.13217}
\showURL{%
\tempurl}


\bibitem[Pitoura et~al\mbox{.}(2018)]%
        {Pitoura2017}
\bibfield{author}{\bibinfo{person}{Evaggelia Pitoura}, \bibinfo{person}{Panayiotis Tsaparas}, \bibinfo{person}{Giorgos Flouris}, \bibinfo{person}{Irini Fundulaki}, \bibinfo{person}{Panagiotis Papadakos}, \bibinfo{person}{Serge Abiteboul}, {and} \bibinfo{person}{Gerhard Weikum}.} \bibinfo{year}{2018}\natexlab{}.
\newblock \showarticletitle{On Measuring Bias in Online Information}.
\newblock \bibinfo{journal}{\emph{SIGMOD Rec.}} \bibinfo{volume}{46}, \bibinfo{number}{4} (\bibinfo{year}{2018}).
\newblock
\showISSN{0163-5808}
\urldef\tempurl%
\url{https://doi.org/10.1145/3186549.3186553}
\showURL{%
\tempurl}


\bibitem[Raza et~al\mbox{.}(2022)]%
        {Raza2022}
\bibfield{author}{\bibinfo{person}{Shaina Raza}, \bibinfo{person}{Deepak~John Reji}, {and} \bibinfo{person}{Chen Ding}.} \bibinfo{year}{2022}\natexlab{}.
\newblock \showarticletitle{Dbias: Detecting biases and ensuring fairness in news articles}.
\newblock \bibinfo{journal}{\emph{International Journal of Data Science and Analytics}} (\bibinfo{year}{2022}), \bibinfo{pages}{1--21}.
\newblock


\bibitem[Rekabsaz et~al\mbox{.}(2021)]%
        {Rekabsaz2021}
\bibfield{author}{\bibinfo{person}{Navid Rekabsaz}, \bibinfo{person}{Simone Kopeinik}, {and} \bibinfo{person}{Markus Schedl}.} \bibinfo{year}{2021}\natexlab{}.
\newblock \showarticletitle{Societal Biases in Retrieved Contents: Measurement Framework and Adversarial Mitigation for BERT Rankers}. In \bibinfo{booktitle}{\emph{Proceedings of the 44th International ACM SIGIR Conference on Research and Development in Information Retrieval (SIGIR '21)}}. \bibinfo{publisher}{ACM}, \bibinfo{address}{Virtual Event, Canada}.
\newblock
\href{https://doi.org/10.1145/3404835.3462949}{doi:\nolinkurl{10.1145/3404835.3462949}}


\bibitem[Singh and Joachims(2019)]%
        {Singh2019}
\bibfield{author}{\bibinfo{person}{Ashudeep Singh} {and} \bibinfo{person}{Thorsten Joachims}.} \bibinfo{year}{2019}\natexlab{}.
\newblock \showarticletitle{Policy Learning for Fairness in Ranking}.
\newblock \bibinfo{journal}{\emph{Advances in neural information processing systems}}  \bibinfo{volume}{32} (\bibinfo{year}{2019}).
\newblock


\bibitem[Singh and Ngu(2025)]%
        {Singh2025}
\bibfield{author}{\bibinfo{person}{Karanbir Singh} {and} \bibinfo{person}{William Ngu}.} \bibinfo{year}{2025}\natexlab{}.
\newblock \bibinfo{title}{Bias-Aware Agent: Enhancing Fairness in AI-Driven Knowledge Retrieval}.
\newblock
\urldef\tempurl%
\url{https://arxiv.org/abs/2503.21237}
\showURL{%
\tempurl}


\bibitem[Spinde et~al\mbox{.}(2021a)]%
        {Spinde2021f}
\bibfield{author}{\bibinfo{person}{Timo Spinde}, \bibinfo{person}{Manuel Plank}, \bibinfo{person}{Jan-David Krieger}, \bibinfo{person}{Terry Ruas}, \bibinfo{person}{Bela Gipp}, {and} \bibinfo{person}{Akiko Aizawa}.} \bibinfo{year}{2021}\natexlab{a}.
\newblock \showarticletitle{Neural Media Bias Detection Using Distant Supervision With {BABE} - Bias Annotations By Experts}. In \bibinfo{booktitle}{\emph{Findings of the Association for Computational Linguistics: EMNLP 2021}}. \bibinfo{publisher}{Association for Computational Linguistics}, \bibinfo{address}{Punta Cana, Dominican Republic}, \bibinfo{pages}{1166--1177}.
\newblock
\href{https://doi.org/10.18653/v1/2021.findings-emnlp.101}{doi:\nolinkurl{10.18653/v1/2021.findings-emnlp.101}}


\bibitem[Spinde et~al\mbox{.}(2021b)]%
        {spinde2021mbic}
\bibfield{author}{\bibinfo{person}{Timo Spinde}, \bibinfo{person}{Lada Rudnitckaia}, \bibinfo{person}{Kanishka Sinha}, \bibinfo{person}{Felix Hamborg}, \bibinfo{person}{Bela Gipp}, {and} \bibinfo{person}{Karsten Donnay}.} \bibinfo{year}{2021}\natexlab{b}.
\newblock \showarticletitle{MBIC--A Media Bias Annotation Dataset Including Annotator Characteristics}.
\newblock \bibinfo{journal}{\emph{arXiv preprint arXiv:2105.11910}} (\bibinfo{year}{2021}).
\newblock


\bibitem[Xu et~al\mbox{.}(2025)]%
        {Xu2025}
\bibfield{author}{\bibinfo{person}{Zhenjie Xu}, \bibinfo{person}{Wenqing Chen}, \bibinfo{person}{Yi Tang}, \bibinfo{person}{Xuanying Li}, \bibinfo{person}{Cheng Hu}, \bibinfo{person}{Zhixuan Chu}, \bibinfo{person}{Kui Ren}, \bibinfo{person}{Zibin Zheng}, {and} \bibinfo{person}{Zhichao Lu}.} \bibinfo{year}{2025}\natexlab{}.
\newblock \bibinfo{title}{Mitigating Social Bias in Large Language Models: A Multi-Objective Approach within a Multi-Agent Framework}.
\newblock
\showeprint[arxiv]{2412.15504}~[cs.CL]
\urldef\tempurl%
\url{https://arxiv.org/abs/2412.15504}
\showURL{%
\tempurl}


\bibitem[Yang and Stoyanovich(2017)]%
        {Yang2017}
\bibfield{author}{\bibinfo{person}{Ke Yang} {and} \bibinfo{person}{Julia Stoyanovich}.} \bibinfo{year}{2017}\natexlab{}.
\newblock \showarticletitle{Measuring fairness in ranked outputs}. In \bibinfo{booktitle}{\emph{Proceedings of Conference on Scientific and Statistical Database Management}}. \bibinfo{pages}{1--6}.
\newblock


\bibitem[Zehlike et~al\mbox{.}(2017)]%
        {Zehlike2017}
\bibfield{author}{\bibinfo{person}{Meike Zehlike}, \bibinfo{person}{Francesco Bonchi}, \bibinfo{person}{Carlos Castillo}, \bibinfo{person}{Sara Hajian}, \bibinfo{person}{Mohamed Megahed}, {and} \bibinfo{person}{Ricardo Baeza-Yates}.} \bibinfo{year}{2017}\natexlab{}.
\newblock \showarticletitle{Fa*ir: A fair top-k ranking algorithm}. In \bibinfo{booktitle}{\emph{Proceedings of the 2017 ACM on Conference on Information and Knowledge Management}}. \bibinfo{pages}{1569--1578}.
\newblock


\bibitem[Zehlike and Castillo(2020)]%
        {Zehlike2020}
\bibfield{author}{\bibinfo{person}{Meike Zehlike} {and} \bibinfo{person}{Carlos Castillo}.} \bibinfo{year}{2020}\natexlab{}.
\newblock \showarticletitle{Reducing disparate exposure in ranking: A learning to rank approach}. In \bibinfo{booktitle}{\emph{Proceedings of The Web Conference}}. \bibinfo{pages}{2849--2855}.
\newblock


\bibitem[Zhang et~al\mbox{.}(2025)]%
        {Zhang2025}
\bibfield{author}{\bibinfo{person}{Zefeng Zhang}, \bibinfo{person}{Hengzhu Tang}, \bibinfo{person}{Jiawei Sheng}, \bibinfo{person}{Zhenyu Zhang}, \bibinfo{person}{Yiming Ren}, \bibinfo{person}{Zhenyang Li}, \bibinfo{person}{Dawei Yin}, \bibinfo{person}{Duohe Ma}, {and} \bibinfo{person}{Tingwen Liu}.} \bibinfo{year}{2025}\natexlab{}.
\newblock \bibinfo{title}{Debiasing Multimodal Large Language Models via Noise-Aware Preference Optimization}.
\newblock
\showeprint[arxiv]{2503.17928}~[cs.CV]
\urldef\tempurl%
\url{https://arxiv.org/abs/2503.17928}
\showURL{%
\tempurl}


\end{thebibliography}
